\documentclass{Interspeech}

\title{Representation of perceived prosodic similarity of conversational feedback}

\author[affiliation={1}]{Livia}{Qian}
\author[affiliation={1}]{Carol}{Figueroa}
\author[affiliation={1}]{Gabriel}{Skantze}

\affiliation{}{KTH Royal Institute of Technology}{Sweden}
\email{liviaq@kth.se, carol.figueroa@etu.univ-amu.fr, skantze@kth.se}

\usepackage{comment}
\usepackage{subfigure}

\begin{document}

\maketitle

\begin{abstract}
    Vocal feedback (e.g., `mhm', `yeah', `okay') is an important component of spoken dialogue and is crucial to ensuring common ground in conversational systems. The exact meaning of such feedback is conveyed through both lexical and prosodic form. In this work, we investigate the perceived prosodic similarity of vocal feedback with the same lexical form, and to what extent existing speech representations reflect such similarities. A triadic comparison task with recruited participants is used to measure perceived similarity of feedback responses taken from two different datasets. We find that spectral and self-supervised speech representations encode prosody better than extracted pitch features, especially in the case of feedback from the same speaker. We also find that it is possible to further condense and align the representations to human perception through contrastive learning.
\end{abstract}


\section{Introduction}

Feedback is a natural part of human conversations, typically expressed as brief listener reactions to what the speaker said. Feedback is multimodal; it can be expressed with e.g., gaze, gestures, head nods and vocally~\cite{poggi2007mind, bevacqua2010multimodal, morency2010probabilistic}. Feedback responses can, for example, signal attention, comprehension, and certainty~\cite{10.1093/applin/19.2.204}. Vocal feedback can take on many forms (e.g., ``right", ``uh-huh", ``yes", ``what?", ``pardon?", ``seriously?", ``alright"). Apart from lexical semantics, the same lexical form can convey different meanings through different prosodic realizations (compare ``yeah...", ``yeah?", and ``yeah!").

To make spoken dialogue systems sound more natural and establish grounding more effectively, it is crucial to investigate the different characteristics of feedback. Previous research has addressed the problems of feedback detection and production through various approaches. Rule-based methods leverage prosodic features such as duration, pitch, intensity, pause length, and changes in $f_0$, particularly in so-called \textit{backchannels}~\cite{Yngve70}, a major subgroup entailing brief, non-interrupting interjections such as ``yeah" and ``okay" \cite{koiso1998analysis, cathcart2003shallow, solorio2006prosodic, truong2010rule}. With recent advances in machine learning, the focus has shifted to self-supervised speech representations that encode diverse aspects of speech~\cite{lin2023utility, chen2021speech, wang2021fine}.

Although the lexical form of feedback can be intuitively modeled using text and language models, prosody is more complex, considering that it is influenced by factors such as speakers' base pitch, accent, and voice quality. In this work, we focus exclusively on the prosodic aspect of feedback and investigate the perceived similarity of feedback responses that have the same lexical form. Our work aims to create \textbf{continuous}, \textbf{low-dimensional}, and \textbf{speaker-independent} prosodic representations that align with human perception of prosodic similarity. By using a continuous representation (a vector space), we allow for nuanced representations where similarity can be measured using distance metrics. Although general speech embeddings represent many aspects of the speech signal, we investigate whether it is possible to condense the representations into fewer dimensions, focusing only on relevant prosodic aspects. 

First, we conduct a perception study in the form of a \textit{triadic comparison task} (e.g., \cite{kotovsky1996comparison,  rich2002alzheimer, lee2016new}) to collect data on how humans contrast English-language feedback in terms of prosody. Triadic comparison tasks present three stimuli per trial, asking participants to either identify the `odd one out' or select the two most similar stimuli (we used the latter approach). Next, we evaluate how well existing representations align with human perception of prosody with pitch-related, spectral, and speech embedding-based metrics. Finally, we refine and condense these representations to align them with the collected perceptual data.

Our results show that certain existing representations strongly correlate with human perception and that they can be improved with contrastive learning. Potential applications of prosodic representations include speech synthesis, along with downstream classification and recognition tasks, such as emotion detection and intent classification.

\section{Related work}

Numerous perception studies have explored the relationship between prosody and the meaning of feedback \cite{wallers2006effect, stocksmeier2007synthesis, chandler2023semantic, neiberg2013semi, figueroa-etal-2024-mhm}. The authors of \cite{wallers2006effect} examined how synthesized prosodic variations of the Swedish feedback /a/ and /m/ affect the perceived meanings in sentences. Similarly, \cite{stocksmeier2007synthesis} examined participants' perceived meanings of synthesized versions of the German feedback ``ja'', presented in isolation rather than in context. The author of \cite{chandler2023semantic} collected responses from various voice talents who were instructed to say ``yeah'' or ``uh-huh'' with a specific intended meaning. \cite{chandler2023semantic} subsequently recruited additional participants to evaluate their perceived meanings and to determine the correlation between perceived and intended meanings provided by voice talents. Collectively, these studies confirm that variations in prosodic features can significantly influence how listeners interpret and ascribe meaning to short feedback expressions across different languages.


Recent studies have investigated whether self-supervised learning models (SSL) capture prosodic information, which they do \cite{kakouros2023does, lin2023utility, deseyssel23_interspeech, yang2023can}. Two benchmarks, SUPERB-prosody \cite{lin2023utility} and ProsAudit \cite{deseyssel23_interspeech}, have been proposed to evaluate the prosodic information that these models can encode. SUPERB-prosody included three downstream tasks: sentiment analysis, sarcasm detection, and persuasiveness prediction. ProsAudit involved identifying strong versus weak prosodic boundaries and distinguishing between different types of inserted pauses. SSLs have been probed at the layer level for prosodic content \cite{kakouros2023does, lin2023utility, yang2023can}, which we also aimed to investigate.





We are unaware of any studies where SSL representations are used to predict human judgment of prosodic similarity. \cite{ward24_interspeech} is an exception, since they collected human judgments of pragmatic similarity for pairs of utterances that have either 1) different lexical content or 2) the same lexical content but where the prosody is different. The authors correlated the perceived pragmatic similarity of human judgments with various SSL models. However, our work differs in that we focus on prosodic similarity in short feedback utterances, and we refine the representations using perception data.

\section{Data}

We used two existing datasets: \textbf{Fisher} English telephone conversations Part 1\footnote{\url{https://catalog.ldc.upenn.edu/LDC2004T19, https://catalog.ldc.upenn.edu/LDC2004S13}} \cite{cieri-etal-2004-fisher} and the \textbf{FiCa} speech dataset\footnote{\url{https://carolfigphd.github.io/FiCa-speech-dataset/}} \cite{figueroa-etal-2024-mhm}. Fisher consists of $5,850$ dyadic telephone conversations between native speakers of U.S. English, each lasting up to around 10 minutes. We used the Montreal Forced Aligner \footnote{\url{https://montreal-forced-aligner.readthedocs.io/en/latest/}} to obtain word-level transcriptions and extracted relevant feedback instances based on their lexical form (listed in Section \ref{sec:stimuli}). FiCa contains $1,685$ recordings of feedback instances produced by a single female voice actor. They were either reenacted in a studio environment or uttered spontaneously in conversation. The reason for choosing these two datasets was that we wished to examine the effects of contrasting feedback from the same versus different speakers.

\section{Perception study}

\subsection{Stimuli}
\label{sec:stimuli}

Since feedback instances from Fisher were extracted automatically, the corresponding recordings required manual review. For the experiments, we curated 400 samples from FiCa and 500 from Fisher, with Fisher data sourced from 320 unique female speakers and 180 male speakers. Although we aimed to include as many Fisher feedback instances as possible during extraction, variations in the distribution of lexical tokens led to an unequal number of each feedback type. To eliminate the effect of lexical semantics, the recordings were grouped by lexical form. During the experiments, randomly selected triads with the same lexical form were presented as stimuli.


The selected feedback lexical forms and corresponding counts for \textbf{FiCa} were: \textit{`no' (30), `okay' (30), `huh' (26), `oh' (25), `ooh' (25), `wow' (25), `mmm' (20), `yeah' (20), `yes' (20), `hm' (15), `mhm' (15), `uh-huh' (15),`ah' (12), `aww' (12), `absolutely' (10), `gosh' (10), `pardon' (10), `right' (10), `ugh' (10), `uh' (10), `uh-oh' (10), `what' (10),          
`jeez' (8),`yup' (8), `exactly' (7), `goodness' (7)}. In the case of \textbf{Fisher}, they were: \textit{ `ah' (36), `oh' (36), `no' (28), `huh' (27), `yup' (26), `uh' (25), `what' (25), `okay' (25), `ooh' (25), `yeah' (25),  `wow'  (25), `yes' (24), `sure' (21),`absolutely' (20), `pardon' (20),  `really' (20), `mhm' (19), `right' (16), `hm' (11), `exactly' (10), `interesting' (10), `mmm' (10), `ugh' (8), `sorry' (8)           }.


\subsection{Task}
Each subject completed 20 tasks, each consisting of three feedback instances with the same lexical form, and an attention check. They were given the instruction ``\textit{Listen to the clips and choose two that are the most similar to each other}". Participants were asked to ignore potential confounding factors that arise from differences in recording devices, such as audio quality and loudness. For Fisher, they were also instructed to ``\textit{[t]ry to focus on basic vocal similarities and differences as opposed to personal factors like gender and whether someone has a higher or lower-pitched voice by default"}.

From the selected feedback instances (400 from FiCa and 500 from Fisher), we randomly created $1,200$ triadic combinations per dataset. Each combination was presented to three subjects, and only those with unanimous agreement were retained. This filtering was necessary, as the elements in some groupings were highly distinct perceptually, or, conversely, practically indistinguishable, leading to uncertain choices. After removing the triplets for which there was no consensus, 486 data points remained for FiCa and 394 for Fisher.
 
\subsection{Participants}

We recruited and paid native U.S. English speakers through the crowdsourcing platform \textit{Prolific}~\footnote{\url{https://prolific.com}}. All participants self-reported having no hearing impairments and claimed that English was their first and primary language, having spent most of their first 18 years in the United States. We enlisted 180 subjects for both datasets.

\section{Evaluation of existing representations}
\label{sec:evaluating_existing}

\subsection{Metrics and evaluation of the study}

We compared various prosodic representations to assess their alignment with the perceived prosodic similarity of feedback instances. First, we used acoustic features based on pitch values extracted with Parselmouth \footnote{\url{https://parselmouth.readthedocs.io/en/stable/}} \cite{jadoul2018introducing}. These included mean pitch, minimum pitch, maximum pitch, pitch range and voiced length (number of voiced frames). To further analyze pitch contours, we fitted third-order Legendre polynomials (LPs) to the extracted pitch values, in the same way as \cite{o2024hierarchical}. The first three coefficients represent height, slope, and convexity. The contour time stamps were normalized between [-1, 1] before fitting the polynomials. We evaluated these coefficients both individually and as concatenated vectors that could be compared using cosine similarity.

Second, we used embeddings from four different speech foundation models: HuBERT~\cite{hubert}, Whisper's encoder~\cite{whisper}, wav2vec 2.0~\cite{baevski2020wav2vec} and W2v-BERT~\cite{chung2021w2v}~\footnote{Pre-trained Huggingface models: \textit{facebook/hubert-large-ls960-ft, openai/whisper-medium, facebook/wav2vec2-large, facebook/w2v-bert-2.0}.}. We used cosine similarity to check the similarity of the embeddings across all layers. Finally, we also investigated spectrogram-based cosine similarity and spectral convergence loss~\cite{arik2018fast}.

In the case of simple scalar values (such as pitch mean), we compare their absolute difference as a measure of similarity. For vector-based representations (e.g., embeddings), we calculated cosine similarity. Using these measures, we determined which pair among the three potential pairings (A-B, A-C, or B-C) was the most similar for a given representation and evaluated how well this aligned with human judgment (yielding a random baseline of 33.33\%). The results are presented in Table~\ref{tab:results1}.

\begin{table}[th]
  \caption{Agreement of different representations with human perception of prosodic similarity (\%). Agreement is based on the correspondence of the human choices with the relative similarity/distance between pairs. For speech embeddings, we provide the range of cosine similarities across all layers.}
  \label{tab:results1}
  \centering
  \begin{tabular}{ lcc }
    \toprule
    \multicolumn{1}{l}{\textbf{Metric}} &
    \multicolumn{1}{c}{\textbf{FiCa}} & 
    \multicolumn{1}{c}{\textbf{Fisher}}  \\
    \midrule
    mean pitch                          & $46.91$  & $\textbf{48.73}$ \\
    min pitch                           & $38.68$  & $\textbf{45.69}$       \\
    max pitch                    & $41.98$  & $\textbf{47.46}$              \\
    voiced length                      & $\textbf{60.49}$  & $56.35$    \\
    pitch range             & $36.42$  & $\textbf{45.17}$    \\
    height (LP curve)            & $47.12$  & $\textbf{47.72}$ \\
    slope (LP curve)            & $\textbf{41.98}$  & $41.88$ \\
    convexity (LP curve)            & $\textbf{41.36}$  & $41.12$ \\
    LP combined cos. sim.           & $\textbf{44.44}$  & $41.88$ \\
    HuBERT cos. sim.             & $\textbf{60.70 -- 72.63}$  & $44.92 - 70.05$ \\
    Whisper cos. sim.             & $\textbf{51.44 -- 68.11}$  & $49.75 - 60.15$ \\
    wav2vec2 cos. sim.             & $\textbf{56.38 -- 73.25}$  & $38.83 - 65.48$ \\
    W2v-BERT cos. sim.             & $\textbf{43.62 -- 76.75}$  & $36.55 - 73.35$ \\
    spectrogram cos. sim.             & $\textbf{65.64}$  & $62.18$ \\
    spectral convergence            & $\textbf{66.26}$  & $63.71$ \\
    \midrule
    random baseline             & $33.33$  & $33.33$ \\
    \bottomrule
  \end{tabular}
  
\end{table}


Pitch-related metrics performed better for Fisher, suggesting that pitch height is more important when distinguishing feedback between multiple speakers. Since this is somewhat counterintuitive, it indicates that some participants may have judged voice similarity rather than prosody, despite instructions to ignore speaker voice characteristics, raising the question of whether, for example, same-gender feedback instances sound more prosodically similar. LP-based height (approximate mean pitch) seems to be slightly more relevant in Fisher, whereas pitch contour convexity and slope -- i.e., pitch changes -- played a greater role in FiCa, especially their combined effect.

Speech embeddings and spectrograms generally capture feedback prosodic similarity better than pitch features. The greater mismatch between model and human judgments in Fisher suggests that pitch metrics and speech models encompass speaker-specific traits, which may lead to speaker-dependent representations. Among pitch-related metrics, voiced length plays a key role in human perception but is less significant when comparing multiple speakers, possibly because listeners focus more on pitch.



\subsection{Effect of embedding layers}

\begin{figure}
    \centering
    \subfigure[FiCa]{
        \includegraphics[width=0.45\linewidth]{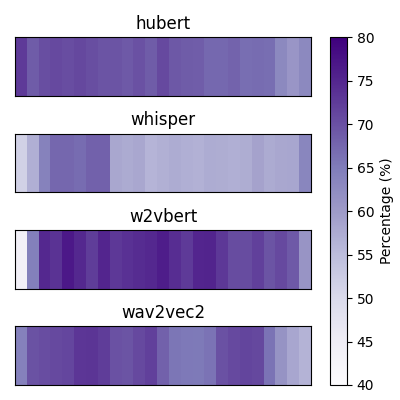}
    }
    \hfill
    \subfigure[Fisher]{
        \includegraphics[width=0.45\linewidth]{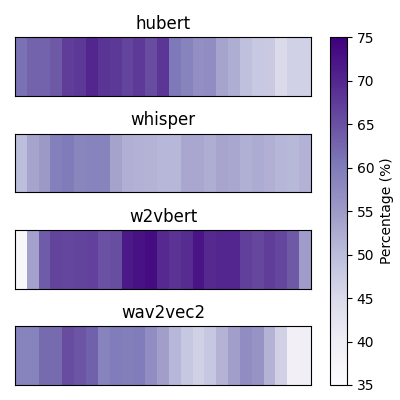}
    }
    \caption{Effect of layers on agreement with people's choices (\%). The layers (0-24) are shown from left to right.}
\label{fig:layers}
\end{figure}

To analyze the role of different layers in the embedding models, we calculated the agreement for each layer in each model, using the variants with 24 layers and a hidden size of 1024. Figure \ref{fig:layers} shows the percentage of agreement between human choices and the cosine similarity of hidden embeddings across all layers (+1 input embedding).

Overall, later layers are less relevant for distinguishing feedback prosody the way humans do. This aligns with \cite{de2024layer} where they found that suprasegmental features like accent and tone are encoded in the middle third of the networks and \cite{pasad2021layer} where they found that shallower layers encode spectral and acoustic features, whereas later layers focus on phonetic, word identity, and word meaning information, in this order.

The figure also shows that the different embeddings show a similar trend for the two datasets. It should be noted that both HuBERT and W2v-BERT show strong overall agreement. However, the last layers, and in some cases, the first, are less representative of prosodic similarity.

\section{Aligning representations with contrastive learning}


As discussed above, the existing representations capture a great amount of information that is not relevant to perceived prosodic similarity. To create more compact and prosodically meaningful representations, we explored projecting them to lower-dimensional spaces and aligning them with human perception using contrastive learning, with the aim of drawing similar representations closer and pushing dissimilar ones apart. We used a triplet loss~\cite{balntas2016learning} with a margin of $0.5$ to train the models on the triplets shown to participants. The learned representations were evaluated in a similar fashion as the original embeddings in Section \ref{sec:evaluating_existing}.

The definition of the loss is:
$$ L(a, p, n) = \max\big\{d(a_i ,p_i) - d(a_i, n_i) + margin, 0\big\} $$
where
$$d(x_i, y_i) = || x_i - y_i||_2$$
and $a$, $p$ and $n$ are a batch of anchors, positive samples and negative samples, respectively. Given a triplet with one positive and two negative connections, one of the elements of the positive connection is always selected at random as the anchor and the other as the positive sample.

We trained the models using the collected perceptual data. We applied 5-fold cross-validation with a holdout test set using a 80-20 split. We investigated whether both pitch contour parameters and speech representation embeddings could be used as input features.

\begin{figure}
    \centering
    \subfigure[FiCa]{
        \includegraphics[width=0.45\linewidth]{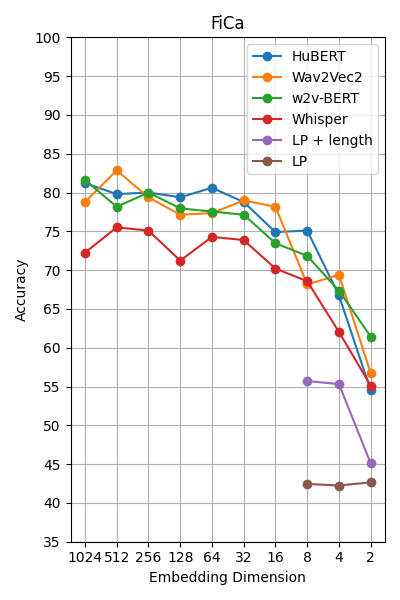}
        \label{fig:model1_1}
    }
    \hfill
    \subfigure[Fisher]{
        \includegraphics[width=0.45\linewidth]{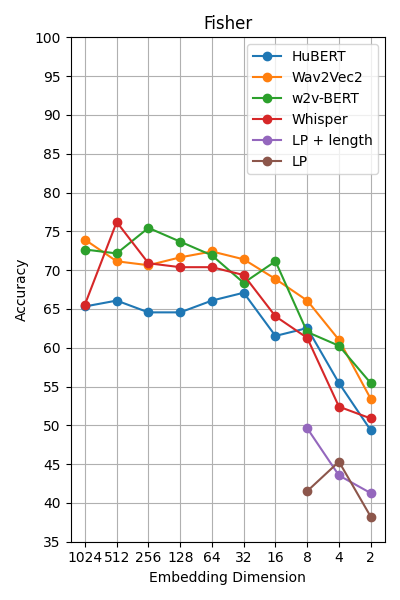}
        \label{fig:model1_2}
    }
    \caption{Test accuracy of the models trained on speech embeddings with different latent sizes, as expressed with cosine similarity (\%). LP + length is the concatenation of the first three LP coefficients and the voiced length, while LP is that of the three coefficients.}
\label{fig:model1}
\end{figure}

For pitch representations, we only used LP coefficients and voiced length. For embeddings, we projected 1024-dimensional vectors from all layers into smaller (or equal-dimensional) spaces using linear projection. We evaluated prosodic expressivity in latent spaces of size $2^n, n \in [1, 2, .., 10]$. LP and voiced length projection, having an input dimension of 3 and 1, respectively, do not necessitate higher dimensional latent spaces; therefore, we assessed them in latent spaces of sizes 2, 4, and 8. The results are shown in Figure \ref{fig:model1}.



The learned embeddings can outperform the original embeddings in higher dimensions, e.g., \textbf{81.22\%} (FiCa, HuBERT with dimension 1024) versus \textbf{72.63\%} (FiCa, pre-trained HuBERT with dimension 1024, as shown in Table \ref{tab:results1}), or \textbf{71.14\%} (Fisher, wav2vec 2.0 with dimension 512) versus \textbf{65.48\%} (Fisher, pre-trained wav2vec 2.0 with dimension 1024). They can also preserve performance in lower dimensions, down to approximately 16, e.g., \textbf{74.90\%} (FiCa, HuBERT with dimension 16) versus \textbf{72.63\%} (FiCa, pre-trained HuBERT with dimension 1024) or \textbf{68.86\%} (Fisher, wav2vec 2.0 with dimension 16) versus \textbf{65.48\%} (Fisher, pre-trained wav2vec 2.0 with dimension 1024). The accuracy decreases slowly after that, but still exceeds the random baseline of 33.33\%.

Although Figure \ref{fig:model1_1} shows that the four embedding types have a relative performance similar to that of the pre-trained embeddings (shown in Table \ref{tab:results1}), Figure \ref{fig:model1_2} shows slightly different trends. HuBERT struggles more with encoding prosodic information from Fisher while Whisper performs comparably to the remaining two embedding types. The reason for this needs to be further investigated, especially in the context of speaker-dependence which mainly constitutes the difference between the two datasets.

As for the Legendre polynomials and pitch length, they do not or barely exceed the performance of their counterparts in Table~\ref{tab:results1} (which indicates 60.49\% and 56.35\% for voiced length, and 44.44\% and 41.88\% for combined LP cosine similarity). However, it is noticeable that LP combined with voiced length is significantly better than LP alone. This marks the duration as a prosodic component that carries meaning.

\section{Discussion and Conclusion}

In this work, we conducted a perception study to analyze how the prosodic perception of conversational feedback aligns with pitch metrics, spectral information, and self-supervised speech embeddings. We then used contrastive learning to condense pre-trained speech representations into prosodically relevant representations. In all cases, we condition on lexical forms to eliminate the effects of lexical semantics.

When it comes to pitch-related features, Legendre polynomials (representing the height, slope, and convexity of the pitch contour) combined with voiced length outperform LPs alone, highlighting the importance of both duration and pitch as relevant prosodic components. In this work, we have not included intensity as a parameter (mainly because it is hard to normalize properly for the perception experiment), but this should also be investigated in future work. 

Speech embeddings using foundation models effectively encode prosodic information, and they do so much better than the basic pitch features used here. While pitch features have the advantage that they are low-dimensional, we show that it is possible to down-project speech embeddings into fewer dimensions (as low as 16, but depending on use-case, even 8 or 4), using perceptual data and contrastive learning. Through this process, the projected embeddings can match or even surpass the performance of the pre-trained speech model embeddings. Our results indicate that contrastive learning based on pairwise human perception is well-suited for obtaining lower-dimensional and more expressive representations in continuous spaces. This method could very well extend to represent prosody of longer utterances as well, and this should be an interesting question for future work. 

We find that prosodic information is encoded primarily in the middle layers in at least four speech foundation models (HuBERT, Whisper, W2v-BERT and wav2vec 2.0) which matches the findings of \cite{de2024layer}. Whisper performed worse, likely due to its ASR-focused training objective, unlike the rest, which are self-supervised models. HuBERT and W2v-BERT provide fairly good representations in general, with the latter displaying a high prosodic content across most layers.

The agreement between embeddings and human perception is lower for Fisher, suggesting that humans rely on different cues when distinguishing feedback from multiple speakers. Pitch height and range play a larger role in differentiating prosody between speakers, while pitch contours are more relevant for feedback from the same speaker. This highlights the need to address speaker independence in prosodic representations.

Representing feedback from multiple speakers appears to be more challenging, as pitch range may subconsciously act as a shortcut for people to cluster similar voices. Additionally, some discrepancies could result from participants not fully engaging with the task, misunderstanding instructions, or overlooking details despite attention checks.

Beyond the presence of multiple speakers, recording quality may also play a crucial role. Fisher recordings often contain background static, which can negatively impact speech embeddings and other metrics. Additionally, Parselmouth occasionally produces poor pitch extraction for Fisher samples. Unlike studio-recorded feedback, Fisher's fully spontaneous feedback lacks the controlled articulation found in FiCa. Further investigation is needed to assess the accuracy of the pitch contours and the influence of background noise.

Our results suggest a promising direction for developing compact prosodic representations of feedback. Another avenue for future work is to explore gender-based differences, given the significant impact of speaker variation on pitch perception. We also plan to apply the learned representations to downstream tasks such as feedback function classification and emotion detection, as well as to improve feedback synthesis and controllability.

\section{Acknowledgements}

This work was partially supported by the Wallenberg AI, Autonomous Systems and Software Program (WASP) funded by the Knut and Alice Wallenberg Foundation, as well as Riksbankens Jubileumsfond (RJ) P20-0484, and Swedish Research Council (VR) 2020-03812.

\bibliographystyle{IEEEtran}
\bibliography{main}

\end{document}